\documentclass[acmtog,nonacm]{acmart}

\usepackage{microtype}
\usepackage{graphicx}
\usepackage{subcaption}
\usepackage{mathtools}
\usepackage[capitalize,noabbrev]{cleveref}

\graphicspath{{./}}
\setcopyright{none}
\authorsaddresses{}

\begin{document}

\title[LCG]{LCG: Long-Context Consistent Image Generation with Sparse Relational Attention}

\author{Zihao Wang}
\authornote{Equal contribution.}
\email{rex.wangzihao@gmail.com}
\affiliation{\institution{Huazhong University of Science and Technology}\country{China}}

\author{Yijia Xu}
\authornotemark[1]
\email{2301213309@stu.pku.edu.cn}
\affiliation{\institution{Peking University}\country{China}}

\author{Haoze Zheng}
\email{hzhengay@connect.ust.hk}
\affiliation{\institution{Hong Kong University of Science and Technology}\country{Hong Kong}}

\author{Xuran Ma}
\email{xmacb@connect.ust.hk}
\affiliation{\institution{Hong Kong University of Science and Technology}\country{Hong Kong}}

\author{Haokun Gui}
\email{hgui@connect.ust.hk}
\affiliation{\institution{Hong Kong University of Science and Technology}\country{Hong Kong}}

\author{Harry Yang}
\authornote{Corresponding author.}
\email{yangharry@ust.hk}
\affiliation{\institution{Hong Kong University of Science and Technology}\country{Hong Kong}}

\renewcommand{\shortauthors}{Wang et al.}

\begin{abstract} 

Recent image generation models achieve impressive quality in single-image synthesis, but often fail to maintain consistency across sequential outputs, as required in comics, storyboards, and visual narratives. We propose Long-Context Generation (LCG), a framework for long-context multi-image text-to-image generation, to improve consistency and scalability in long-context multi-image generation. LCG employs the Sparse Relational Attention (SRA) mechanism to selectively attend to core features across extended visual contexts, ensuring that the propagation of semantic and layout information remains computationally tractable.
To enforce semantic alignment, we introduce the Routing Consistency Constraint (RCC), which leverages identity-aware masks to align structural patterns across generation branches, effectively mitigating drift in appearance even in complex multi-character scenes.
To support training and evaluation in this setting, we construct the Long-Context Consistency Dataset (LCCD), a large-scale synthetic dataset comprising character-centric multi-image sequences spanning varied situational contexts. LCCD contains 600K training sequences and a separate 1K test set, with each sequence containing 6 to 20 images. The experiments demonstrate that LCG outperforms the compared baselines in prompt alignment and character consistency for long-context image generation, including multi-character scenes.
\end{abstract}

\maketitle

\section{Introduction}
\label{sec:intro}

Generative models have made remarkable progress in producing high-quality single images, yet many real-world creative scenarios require models to reason over long sequences or collections of related images. Tasks such as film key-frame design, multi-view editing, comics, and visual storytelling demand not only high-fidelity generation for individual images, but also semantic, structural, and stylistic coherence across the entire sequence. In this work, we focus on long-context multi-image text-to-image generation: given a sequence of scene-specific text prompts, the model should jointly synthesize a corresponding sequence of images that follow their individual prompts while preserving consistent character identity, role assignment, and visual continuity.

Existing methods~\cite{mou2025dreamo,dinkevich2025story2board,ma2024storynizorconsistentstorygeneration,luo2025objectisolatedattentionconsistent,li2025iccustomdiverseimagecustomization} often struggle in this setting. Generating panels independently ignores cross-panel evidence, while one-by-one sequential generation can propagate early identity drift or role mismatches to later panels. Dense attention across all panels provides a direct way to exchange information, but its quadratic memory scaling quickly exceeds practical memory budgets as the number of panels grows~\cite{xiao2025captaincinemashortmovie,wu2025omnigen2explorationadvancedmultimodal,wu2025lightgenefficientimagegeneration,ma2025modelrevealscacheprofilingbased,yuan2025nativesparseattentionhardwarealigned}. As a result, developing models that can efficiently exploit long-range visual context while preserving global coherence remains an open challenge~\cite{rombach2022high,peebles2023scalable,podell2023sdxl,wang2024oneactorconsistentcharactergeneration,chen2025unireal}.

To address this challenge, we introduce Long-Context Generation (LCG), a diffusion-based framework for long-context consistent image generation. LCG assigns each prompt to a parallel generation branch and denoises all branches jointly, so that different panels can exchange semantic and identity-related evidence throughout the generation process instead of relying on fixed previously generated outputs. To jointly target scalability and consistency, LCG reformulates attention into a Sparse Relational Attention (SRA) mechanism, allowing each generation branch to attend only to the most relevant regions from other branches. This selective information flow enables long-context computation within practical memory budgets and helps propagate semantic and layout cues across distant images, ensuring global coherence. 
To further promote semantic correspondence and identity preservation, we introduce the Routing Consistency Constraint (RCC) loss. This mechanism leverages identity-aware masks to enforce semantic and structural correspondence across parallel generation branches, reducing drift in appearance and composition even in complex multi-subject interactions without requiring dense supervision.

Furthermore, we construct the Long-Context Consistency Dataset (LCCD), a controlled synthetic dataset for training and evaluating models under extended contextual conditions. LCCD provides 600K training sequences, split into 300K single-character and 300K multi-character sequences, together with a separate 1K test set containing 500 single-character and 500 multi-character sequences. Each sequence contains 6 to 20 images, a significant expansion over the 2-4 frame span of existing datasets. To support reliable training and evaluation under controlled protocols, LCCD is verified through a structured filtering pipeline that discards inconsistent samples.
Extensive experiments show that LCG substantially improves character and scene consistency, as well as overall visual fidelity, while maintaining practical computational cost.
Our contributions are summarized as follows:
\begin{itemize}
    \item We propose \textbf{Long-Context Generation (LCG)}, a framework for long-context multi-image text-to-image generation. LCG improves cross-panel consistency and scalability by enabling sparse information exchange among parallel generation branches.
    \item We introduce \textbf{Sparse Relational Attention (SRA)}, a sparse cross-branch routing module that combines compact semantic selection with fine-grained relational token retrieval across jointly denoised generation branches. We further develop the \textbf{Routing Consistency Constraint (RCC)}, which converts identity-aware masks into token-level routing targets to encourage consistent character identity, scene layout, and structural correspondence throughout the sequence.
    \item We construct the \textbf{Long-Context Consistency Dataset (LCCD)}, a large-scale synthetic dataset featuring character-centric images across varied situational contexts. Its separate test set supports systematic evaluation of multi-subject generation under long-context settings. Our evaluations demonstrate that LCG significantly improves identity preservation and consistency.
\end{itemize}

\begin{figure*}[t]
  \centering
  \includegraphics[width=0.87\textwidth]{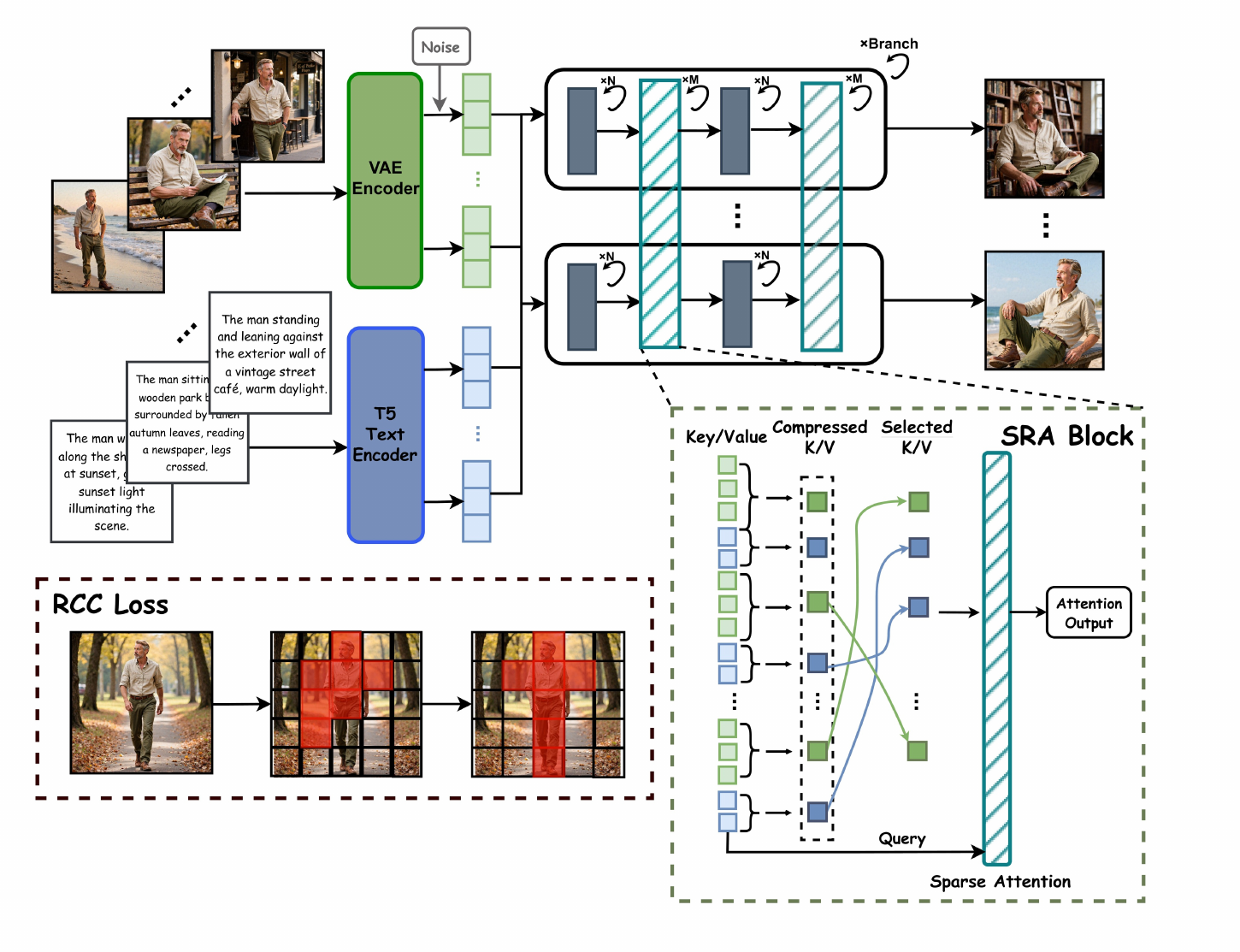}
  \caption{Overview of the proposed LCG pipeline. 
  The model integrates Sparse Relational Attention (SRA) and Routing Consistency Constraint (RCC) to achieve consistent multi-image generation.}
  \Description{A pipeline diagram illustrating the LCG framework. Multiple generation branches exchange information through Sparse Relational Attention, while Routing Consistency Constraint supervises cross-branch alignment using identity-aware masks.}
  \label{fig:pipeline}
\end{figure*}

\section{Related work}

\subsection{Controllable Text-to-Image Generation}
Text-to-image diffusion models operating in a learned latent space have become the mainstream paradigm for high-fidelity conditional image synthesis. Early works such as Latent Diffusion Model (LDM)~\cite{rombach2022high}, Diffusion Transformer (DiT)~\cite{peebles2022dit}, and Stable Diffusion XL~\cite{podell2023sdxl} demonstrated that combining compact latent representations with transformer-based denoisers enables efficient and expressive image generation guided by natural language prompts. Subsequent research has focused on enhancing controllability and structural alignment between textual and visual domains. ControlNet~\cite{zhang2023adding} introduces external conditioning branches to integrate structural cues such as edge maps, poses, or depth maps, while T2I-Adapter~\cite{mou2024t2i} employs lightweight adapters to inject control signals into frozen diffusion backbones. Other variants like IP-Adapter~\cite{ye2023ip} and PhotoMaker~\cite{li2023photomaker} extend this idea to subject-aware conditioning, allowing reference images to guide visual identity during generation.

In parallel, methods such as DreamBooth~\cite{ruiz2023dreambooth}, Custom Diffusion~\cite{kumari2023multi}, and Textual Inversion~\cite{gal2022image} personalize text-to-image models through fine-tuning or embedding optimization, enabling instance-level control over appearance. Despite these advances, most controllable diffusion frameworks focus on single-image customization or subject conditioning, rather than joint long-context multi-image text-to-image synthesis. Our Long-Context Generation (LCG) addresses this limitation by employing a sparse attention mechanism to enable controllable generation over extended multi-image contexts, achieving improved semantic and identity consistency.

\subsection{Storyboard Generation}

Storyboard generation extends text-to-image diffusion toward multi-panel synthesis, where visual coherence, narrative flow, and character consistency must be simultaneously maintained. Unlike standard text-to-image generation, which focuses on individual images, storyboard generation aims to create temporally and semantically linked sequences that depict a narrative or visual story. Recent studies such as StoryDiffusion~\cite{zhou2024storydiffusion} introduce consistency-aware attention mechanisms and semantic motion predictors to preserve temporal coherence across panels. Story2Board~\cite{dinkevich2025story2board} is a training-free storyboard generation method that uses Latent Panel Anchoring (LPA) and Reciprocal Attention Value Mixing (RAVM) to improve cross-panel consistency. StoryGen~\cite{liu2024intelligent} leverages an autoregressive vision-language generation pipeline, progressively producing coherent visual narratives conditioned on preceding frames. DreamStory~\cite{he2025dreamstory} adopts a language-model-driven prompt decomposition strategy coupled with a multi-subject diffusion backbone to maintain inter-character relationships and story dynamics.

Other approaches, including IC-LoRA~\cite{huang2024context} and OminiControl~\cite{tan2025ominicontrol}, explore lightweight adaptation or spatial layout conditioning to improve frame-to-frame consistency and stylistic continuity. UNO~\cite{wu2025less} proposes a multi-image conditioned generation framework that incorporates progressive cross-modal alignment and position encoding to handle multiple reference images, supporting consistent synthesis across several panels. Methods like InstantID~\cite{wang2024instantid} extend identity-based conditioning by using reference portraits and pose guidance to preserve facial characteristics across generated results. These methods mainly target reference- or subject-conditioned generation protocols, whereas our main evaluation setting takes only a sequence of text prompts as input and jointly synthesizes the corresponding image sequence. Our LCG mitigates long-context drift by introducing sparse attention for efficient long-range dependency modeling, maintaining coherence and consistency in extended multi-panel generation.

\section{Method}
\label{sec:Method}

Our main task is long-context multi-image text-to-image generation. Given a sequence of scene-specific natural language prompts $\mathcal{P}=\{p_i\}_{i=1}^{N}$, where each prompt describes the desired subject appearance, action, scene, and camera condition for one panel, the goal is to jointly generate an image sequence $\mathcal{X}=\{x_i\}_{i=1}^{N}$ such that each image aligns with its corresponding prompt while the whole sequence preserves character identity, role assignment, and visual continuity. An overview of our method is shown in \cref{fig:pipeline}. LCG assigns each prompt to a parallel generation branch and denoises all branches jointly from noise. To ensure consistency under long contexts, we introduce two complementary mechanisms: the Sparse Relational Attention (SRA) module and the Routing Consistency Constraint (RCC). SRA implements a multi-branch attention mechanism in which each branch selectively attends to the most relevant tokens from other branches. RCC enhances visual consistency across generation branches through a mask-guided alignment loss that directs attention toward semantically corresponding regions, improving identity fidelity and overall structural coherence.

\subsection{Long-Context Consistent Multi-Image Generation}

We first introduce Sparse Relational Attention (SRA), which supports long-context consistency by allowing each generation branch to selectively retrieve identity-related and semantic evidence from other jointly denoised branches.

The key challenge is that cross-branch attention is both expensive and noisy in long-context generation. Dense attention computes all token interactions across all panels, which quickly becomes impractical as the number of branches grows. More importantly, most cross-branch token pairs are not semantically useful: a query describing one character, object, or scene region may attend to unrelated tokens from other panels, causing identity drift, role confusion, or layout interference. SRA addresses this challenge by formulating long-context interaction as query-conditioned evidence routing among jointly denoised branches. It first builds compact block-level summaries over all branches to identify globally relevant regions, and then retrieves fine-grained local and relational tokens from the selected candidates. Each sparse route therefore specifies which identity, layout, and scene evidence can be exchanged across panels. This concentrates cross-branch communication on semantically plausible relations while keeping long-context generation computationally tractable.

\paragraph{Notation and Setup.}
Let $N$ denote the number of parallel generation branches, $L$ the token sequence length per branch, and $M$ the total number of transformer layers. Each branch possesses its own query, key, and value matrices, denoted as $Q_b, K_b, V_b \in \mathbb{R}^{L \times D}$ ($b \in \{1, \dots, N\}$). Concatenating across branches yields the combined representations:
\begin{equation}
K_{1:N} = [K_1; \dots; K_N], \quad V_{1:N} = [V_1; \dots; V_N].
\end{equation}

\paragraph{Coarse Semantic Compression.}
To build a compact semantic overview of all branches, we compress each sequence into block-level representations. $\phi_{cmp}$ is a linear projection followed by a learnable aggregation operator, and $w$ is the window size:
\begin{align}
\tilde{K}^{cmp}_b = \phi_{cmp}(K_b[i:i+w-1]), \\
\tilde{V}^{cmp}_b = \phi_{cmp}(V_b[i:i+w-1]),
\end{align}
These compressed features form a global semantic map:
\begin{equation}
\tilde{K}^{cmp} = [\tilde{K}^{cmp}_1; \dots; \tilde{K}^{cmp}_N].
\end{equation}

\paragraph{Local Context Selection.}
Each query $q_i^b \in Q_b$ first interacts with its own branch to retrieve the top-$K_{loc}$ most relevant tokens:
\begin{align}
s^{local}_{ij} &= \frac{q_i^b (k_j^b)^\top}{\sqrt{D}}, \\
S_{i,local}^b &= \text{TopK}(s^{local}_{i:}, K_{loc}),
\end{align}
forming local subsets:
\begin{align}
\tilde{K}^{local}_{i,b} = \{ k_j^b \mid j \in S_{i,local}^b \}, \\ 
\tilde{V}^{local}_{i,b} = \{ v_j^b \mid j \in S_{i,local}^b \}.
\end{align}

\paragraph{Global Relational Selection.}
To incorporate long-range dependencies, each query also attends to the compressed semantic representations from all $N$ branches:
\begin{equation}
p_i^{cmp} = \text{Softmax}\left( \frac{q_i^b (\tilde{K}^{cmp})^\top}{\sqrt{D}} \right).
\end{equation}
The top-$K_g$ relevant blocks are selected:
\begin{equation}
I_i = \text{TopK}(p_i^{cmp}, K_g),
\end{equation}
where $I_i$ denotes the set of selected block indices. The corresponding fine-grained tokens within these selected blocks are retrieved to form global subsets $\tilde{K}^{global}_{i,b}$ and $\tilde{V}^{global}_{i,b}$. Tokens from the same branch can be optionally excluded to strengthen inter-branch reasoning.

\paragraph{Sparse Relational Attention.}
Each query attends over the union of local and global candidates:
\begin{equation}
S_i^b = S_{i,local}^b \cup \Omega(I_i),
\end{equation}
where $\Omega(I_i)$ denotes the set of fine-grained tokens contained within the selected blocks $I_i$. The sparse attention is then computed as:
\begin{align}
\alpha_{ij} &=
\frac{\exp\left(\frac{q_i^b \cdot k_j}{\sqrt{D}}\right)}
{\sum_{j' \in S_i^b} \exp\left(\frac{q_i^b \cdot k_{j'}}{\sqrt{D}}\right)}, \\
o_i^b &= \sum_{j \in S_i^b} \alpha_{ij} v_j.
\end{align}
The outputs $O_b = \{o_i^b\}_{i=1}^L$ are then projected and propagated through the diffusion backbone for subsequent denoising steps.

\paragraph{Complexity Analysis.}
The proposed SRA significantly reduces the resource requirements under long visual contexts by separating coarse block selection from fine-grained sparse aggregation. Full dense attention over all generation branches scales as $\mathcal{O}(N^2L^2D)$. In SRA, the coarse block selection scores compressed block summaries from all branches, which costs approximately $\mathcal{O}(N^2L^2D/w)$ with compression window size $w$. After selecting a fixed number of candidates, SRA restricts each query to only $K_{loc}$ local tokens and $K_{glob}$ global tokens (from $\Omega(I_i)$), yielding a fine-grained sparse attention cost of:
\begin{equation}
\mathcal{O}(NL(K_{loc} + K_{glob})D), \quad K_{loc}, K_{glob} \ll NL.
\end{equation}
This design reduces the dense quadratic interaction through block compression and bounds the fine-grained attention by fixed candidate counts, while preserving selected long-range relational interactions across generation branches.

\paragraph{Branch-Level Temporal Encoding.}
To help the model distinguish multiple simultaneous generation branches, LCG injects a branch-specific identifier into the temporal dimension of the 3D Rotary Position Embedding (3D RoPE). Specifically, for each generation branch $b$ assigned to prompt $p_b$, we assign a fixed unique branch identifier $\tau_b$ and use it when computing the temporal component of RoPE for branch $b$. This branch-level temporal signature provides each branch with an explicit positional identity, helping SRA distinguish different panels and reducing identity ambiguity across long-context generation branches. In the optional user-specified identity setting, clean identity evidence can be injected into selected branches using the same branch-level encoding, but reference images are not required by the main multi-image text-to-image task.

\subsection{Routing Consistency Constraint (RCC)}
SRA makes long-context interaction tractable, but its sparse routes are still selected from model features and can be unstable during training. When different characters share similar clothing, poses, or backgrounds, a branch may route attention to visually similar but semantically incorrect regions in another branch. This failure is especially harmful for multi-image generation because a wrong cross-branch route can repeatedly inject identity or role evidence into the wrong panel. To stabilize this process, we introduce the \textit{Routing Consistency Constraint (RCC)} loss, which regularizes cross-branch attention as semantic routing.

The intuition is that identity-aware masks provide a lightweight correspondence signal across panels. Since each mask channel corresponds to the same character identity throughout a training sequence, tokens assigned to the same character in different branches should form stronger cross-branch routes than unrelated regions. RCC converts these same-character correspondences into normalized routing targets and penalizes deviations between the predicted sparse attention map and the mask-derived correspondence. In this way, SRA determines the candidate communication paths, while RCC encourages these paths to remain semantically grounded, reducing identity and role misrouting in multi-character long-context generation.

Formally, let $D$ denote the hidden dimension. For each layer $\ell \in \{1, \dots, M\}$, we consider a generation branch $b$ and another branch $c \neq b$. Branch $b$ provides the queries to be updated, while branch $c$ provides the keys and values attended by branch $b$. The corresponding row-normalized cross-branch attention map is computed as:
\begin{equation}
A_{b,c}^{(\ell)}
=
\mathrm{Softmax}_{\mathrm{row}}\!\left(
\frac{Q_b^{(\ell)} \bigl(K_c^{(\ell)}\bigr)^{\top}}{\sqrt{D}}
\right)
\in [0,1]^{L\times L}.
\label{eq:rcc_attention}
\end{equation}
where $Q_b^{(\ell)} \in \mathbb{R}^{L \times D}$ and $K_c^{(\ell)} \in \mathbb{R}^{L \times D}$ represent the query matrix of branch $b$ and the key matrix of branch $c$, respectively. We then retain only the top-$k$ strongest entries per query token and renormalize each row:
\begin{equation}
\tilde{A}_{b,c}^{(\ell)}
=
\mathrm{RowNorm}\!\left(
\mathcal{T}_k\!\left(A_{b,c}^{(\ell)}\right)
\right)
\in [0,1]^{L\times L}.
\label{eq:rcc_sparse_attention}
\end{equation}
Here $\mathcal{T}_k(\cdot)$ keeps the top-$k$ strongest attention entries per row and sets the remaining entries to zero. This step yields a structured yet computationally efficient cross-branch attention pattern.

To construct the supervision signal, suppose that a training sequence contains $R$ character identities. Let $M_b^{H\times W\times R}$ and $M_c^{H\times W\times R}$ denote the high-resolution identity-aware mask tensors of branches $b$ and $c$ in the original image space, where the $r$-th channel corresponds to the same character identity across the whole sequence. We downsample them to match the attention resolution using an adaptive pooling operator $D_M(\cdot)$:
\begin{equation}
\begin{split}
M_b &= D_M\bigl(M_b^{H\times W\times R}\bigr)\in[0,1]^{L\times R},\\
M_c &= D_M\bigl(M_c^{H\times W\times R}\bigr)\in[0,1]^{L\times R}.
\end{split}
\end{equation}
Based on the shared character channels, we derive a normalized target routing map that captures the desired same-identity correspondence between branch $b$ and branch $c$:
\begin{equation}
\begin{split}
C_{b,c} &= M_b M_c^{\top}\in[0,1]^{L\times L},\\
\hat{A}_{b,c}^{(\ell)}
&=
\mathrm{RowNorm}\!\left(C_{b,c}\right).
\end{split}
\label{eq:rcc_target}
\end{equation}
Thus, $C_{b,c}[i,j]$ is positive when token $i$ in branch $b$ and token $j$ in branch $c$ belong to the same character identity, which explicitly defines multi-character matching across panels. $\mathrm{RowNorm}(\cdot)$ normalizes rows with positive mass and keeps all-zero rows unchanged. To restrict supervision to character regions, we define a foreground query mask:
\begin{equation}
\begin{split}
f_b &= \min\!\left(M_b\mathbf{1}_R, \mathbf{1}_L\right)\in[0,1]^L,\\
W_b &= f_b\mathbf{1}_L^{\top}\in[0,1]^{L\times L},
\end{split}
\label{eq:rcc_weight}
\end{equation}
where $\mathbf{1}_R$ and $\mathbf{1}_L$ are all-one vectors and the minimum is applied element-wise. To ensure stable supervision across hierarchical representations, the consistency loss is computed layer-wise and averaged over all supervised cross-branch terms.

\begin{equation}
\mathcal{L}_{\mathrm{RCC}}
=
\frac{1}{Z}
\sum_{\ell=1}^{M}
\sum_{b=1}^{N}
\sum_{\substack{c=1\\c\neq b}}^{N}
\Big\|
W_b\odot
\left(
\tilde{A}_{b,c}^{(\ell)}
-
\hat{A}_{b,c}^{(\ell)}
\right)
\Big\|_F^2,
\label{eq:rcc_loss}
\end{equation}
where $\odot$ denotes element-wise masking, $\|\cdot\|_F$ is the Frobenius norm, and $Z=\sum_{\ell=1}^{M}\sum_{b=1}^{N}\sum_{c\neq b}\|W_b\|_1$ normalizes by the number of supervised foreground entries. The model is optimized using a weighted combination of the latent diffusion loss and the consistency constraint: $\mathcal{L}_{total} = \mathcal{L}_{ldm} + \lambda \mathcal{L}_{\mathrm{RCC}}$, where $\lambda$ controls the strength of the alignment. Minimizing $\mathcal{L}_{\mathrm{RCC}}$ encourages the network to produce stable, semantically aligned cross-branch attention flows, which improves multi-branch coherence in long-context multi-image synthesis.

\section{Experiment}
\label{sec:experiment}
We implement LCG on top of the Flux.1-dev backbone. The pre-trained Flux.1-dev weights are frozen, and only the newly introduced SRA modules are optimized with the diffusion objective and the RCC loss. Following the architecture described in Section~\ref{sec:Method}, we uniformly insert three SRA modules within each Double Block and six SRA modules within each Single Block. The coarse semantic compression window is set to $w = 128$, and the global block selection count is set to $K_g = 64$. Additional training hyperparameters are provided in Appendix~\ref{app:implementation}.

Training is conducted in two stages: Stage 1 uses the 300K single-character training sequences, and Stage 2 continues on the 300K multi-character training sequences. The test set contains 1K sequences, split into 500 single-character and 500 multi-character sequences.

Section 4.1 introduces the compared methods, evaluation metrics, and the LCCD dataset. Section 4.2 reports qualitative, quantitative, and human-preference comparisons on long-context multi-image text-to-image generation. Section 4.3 provides ablation studies, including user-specified identity generation, attention design, RCC, and context length. More details are provided in the appendices.

\subsection{Baselines, Metrics and Dataset}
We compare LCG with Flux.1-dev, StoryDiffusion, and Story2Board \cite{dinkevich2025story2board} under the same multi-image text-to-image setting. StoryDiffusion \cite{zhou2024storydiffusion} targets text-based visual story generation and uses Consistent Self-Attention to improve consistency across generated image sequences, while Story2Board is a training-free storyboard generation method based on Latent Panel Anchoring (LPA) and Reciprocal Attention Value Mixing (RAVM).

We evaluate prompt alignment with VQAScore~\cite{lin2024evaluating}, subject consistency with DreamSim similarity~\cite{fu2023dreamsim}, facial identity preservation with FaceSim-Arc~\cite{deng2019arcface}, and visual quality with Aesthetic Score~\cite{schuhmann2022improvedaesthetic}. All metrics are averaged over each generated sequence; implementation details are provided in Appendix~\ref{app:implementation}.
To evaluate our model under extended contextual conditions, we introduce the Long-Context Consistency Dataset (LCCD), a large-scale photorealistic and controllable synthetic dataset for long-context multi-image generation. The dataset comprises 600K training sequences, split into 300K single-character and 300K multi-character sequences, and a separate 1K test set, split into 500 single-character and 500 multi-character sequences. Each sequence contains 6 to 20 semantically coherent prompts and their corresponding images.
The training-set distributions over sequence length and character count are visualized in \cref{fig:lccd_distribution}.

We adopt synthetic construction instead of collecting real movie or web-video frames because identifiable human subjects and copyrighted media can introduce legal and ethical risks related to copyright, privacy, likeness rights, and identity consent. Synthetic generation also allows us to construct controllable and diverse character profiles, scenes, actions, camera views, and multi-character compositions at scale.

We construct LCCD through a structured data generation pipeline. Character identities and attributes are first synthesized using Gemini 3, which provides detailed and diverse role descriptions. Based on these character definitions, we generate scene prompts following a concise compositional template that combines camera and style cues, character appearance, action, environment, and lighting. Each prompt sequence is synthesized using Seedream 4.0. To mitigate identity drift in extended sequences, we employ Gemini 3 to perform automatic consistency screening for each image set, discarding sequences with identity drift, role confusion, severe clothing changes, missing characters, or obvious visual artifacts. Grounded-SAM is utilized for instance-level segmentation to generate character masks for RCC supervision. Detailed implementation steps are provided in Appendix~\ref{app:lccd_details}.

\begin{table}[t]
  \caption{Comparison of methods on prompt alignment, subject consistency, facial identity preservation, and visual quality. Our method achieves the best overall performance.}
  \label{tab:results}
  \centering
  \footnotesize 
  \setlength{\tabcolsep}{6pt} 
  \resizebox{\linewidth}{!}{
  \begin{tabular}{@{}lcccc@{}}
    \toprule
    Method & \begin{tabular}[c]{@{}c@{}}Prompt Alignment\\(VQA Score) $\uparrow$\end{tabular} 
           & \begin{tabular}[c]{@{}c@{}}Subject Consistency\\(DreamSim Sim.) $\uparrow$\end{tabular}
           & \begin{tabular}[c]{@{}c@{}}Identity Preservation\\(FaceSim-Arc) $\uparrow$\end{tabular}
           & \begin{tabular}[c]{@{}c@{}}Visual Quality\\(Aesthetic Score) $\uparrow$\end{tabular} \\
    \midrule
    StoryDiffusion & 0.52 & 0.62 & 0.49 & 0.56 \\
    Story2Board & 0.64 & 0.62 & 0.52 & 0.58 \\
    Flux.1-dev & 0.55 & 0.63 & 0.46 & 0.60 \\
    \textbf{LCG (Ours)} & \textbf{0.71} & \textbf{0.75} & \textbf{0.66} & \textbf{0.62} \\
    \bottomrule
  \end{tabular}
  }
\end{table}

\subsection{Comparison on Long-Context Consistent Image Generation}
\subsubsection{Qualitative Evaluation}
\begin{figure*}[t]
    \centering
    \includegraphics[width=0.88\textwidth]{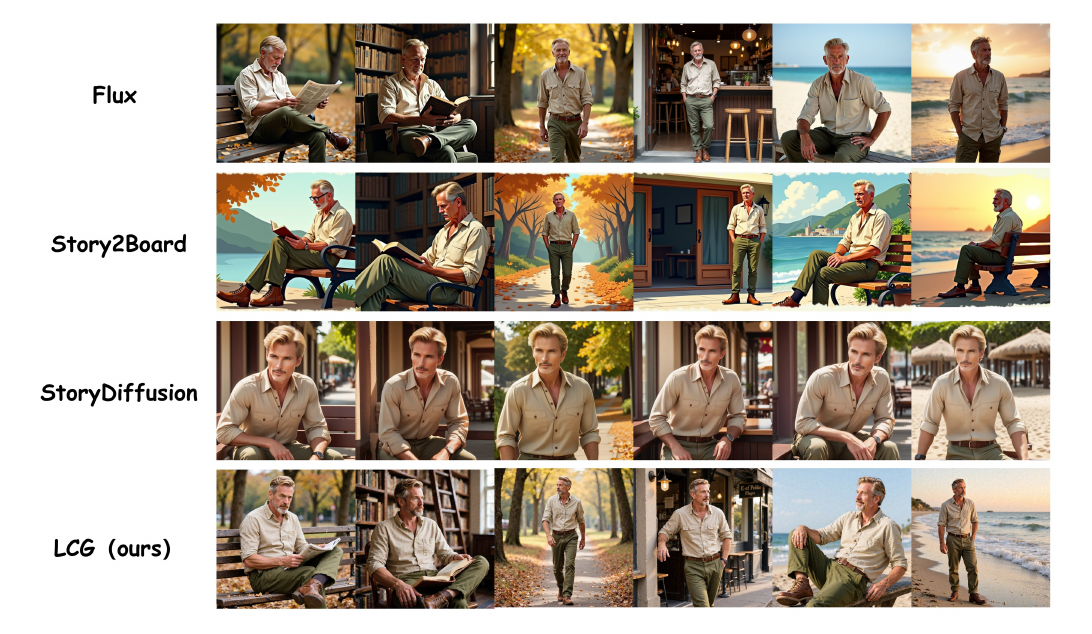}
    \caption{Qualitative comparison on long-context multi-image text-to-image generation. Rows correspond to Flux.1-dev, Story2Board, StoryDiffusion, and LCG. The examples are generated from the same prompt sequence: 
    (1) The old man with grey hair and beard, wearing a light beige shirt and olive green pants, sitting on a wooden bench in an autumn park, reading a newspaper; 
    (2) \ldots{} sitting in a library with floor-to-ceiling bookshelves, holding an open book; 
    (3) \ldots{} walking on a path through an autumn forest with yellow leaves under bright sunlight; 
    (4) \ldots{} leaning against the exterior wall of a vintage street cafe in warm daylight; 
    (5) \ldots{} sitting on a bench by the seaside, looking at the calm ocean; 
    (6) \ldots{} standing on a sandy beach at sunset, waves in the background.}
    \Description{A side-by-side qualitative comparison with four rows corresponding to Flux.1-dev, Story2Board, StoryDiffusion, and LCG. Each row contains six generated panels for the same elderly man prompt sequence across different scenes.}
    \label{fig:main}
\end{figure*}
\Cref{fig:main} shows qualitative comparisons. LCG maintains consistent character appearance across multiple panels while following scene-specific prompts.
\subsubsection{Quantitative Evaluation}
We report quantitative comparisons in Table~\ref{tab:results}. We evaluate prompt alignment with VQAScore, subject consistency with DreamSim similarity, facial identity preservation with FaceSim-Arc, and visual quality with Aesthetic Score. LCG obtains stronger scores on the reported metrics, indicating improved identity preservation, subject consistency, and prompt following.

\subsubsection{User Study}
We randomly sample 50 prompt sequences from the LCCD test set, covering both single-character and multi-character cases. We collect valid annotations from 30 participants. For each sequence, participants compare anonymized and randomly ordered results from Flux.1-dev, StoryDiffusion, Story2Board, and LCG, and rate each result on a 1--5 scale in terms of text alignment, identity consistency, visual quality, and overall preference.

\begin{table}[t]
\centering
\caption{Human evaluation on long-context multi-image text-to-image generation. LCG receives the highest overall preference and stronger human preference for text alignment and identity consistency. Higher scores indicate better performance.}
\footnotesize
\setlength{\tabcolsep}{3pt}
\resizebox{\linewidth}{!}{
\begin{tabular}{lcccc}
\toprule
\textbf{Method} & \textbf{Text Align. $\uparrow$} & \textbf{Identity Consist. $\uparrow$} & \textbf{Visual Quality $\uparrow$} & \textbf{Overall Pref. $\uparrow$} \\
\midrule
Flux.1-dev & 3.34 & 3.03 & \textbf{3.85} & 3.41 \\
StoryDiffusion & 3.26 & 3.38 & 3.47 & 3.37 \\
Story2Board & 3.53 & 3.34 & 3.53 & 3.47 \\
\textbf{LCG (Ours)} & \textbf{4.08} & \textbf{4.18} & 3.82 & \textbf{4.03} \\
\bottomrule
\end{tabular}
}
\label{tab:user_study}
\end{table}
\subsection{Ablation Study}
\subsubsection{User-Specified Identity Generation}
We further evaluate whether LCG can incorporate user-specified identity evidence during long-context multi-image generation. To test this capability, clean identity images are provided to selected generation branches as initialization conditions, while the remaining branches start from noise and are jointly denoised according to their text prompts. Through SRA-based cross-branch interaction, identity evidence from the selected branches can be propagated to the other generated panels, enabling the sequence to preserve the specified identity while still following panel-specific scenes and actions. This experiment examines LCG's ability to support image-based identity conditioning and is not used as the primary evaluation setting.

\begin{figure}[t]
    \centering
    \includegraphics[width=\linewidth]{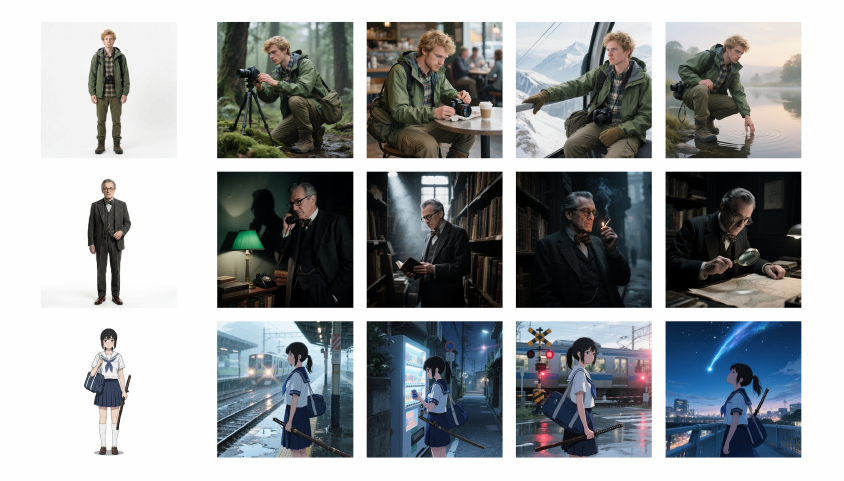}
    \caption{Qualitative examples of user-specified identity generation. The left column shows the clean identity images, and the remaining columns show generated panels. Selected branches are provided with clean identity images instead of pure noise, while the remaining branches are jointly denoised from text prompts. LCG propagates the user-specified identity across the generated sequence and preserves consistent appearance under different scenes and actions.}
    \Description{Qualitative examples of user-specified identity generation, showing clean identity images used as branch-level evidence and multiple generated panels that preserve the specified identity across different prompts.}
    \label{fig:ref}
\end{figure}

As shown in \cref{fig:ref}, LCG generates prompt-specific panels that remain consistent with the provided identity images. This indicates that the joint generation process can incorporate clean identity evidence while maintaining long-context consistency. Additional examples are provided in the appendices.

\subsubsection{Attention Configuration}
Table~\ref{tab:ablation_attention} compares attention configurations in LCG. 
Replacing full attention with the proposed Sparse Relational Attention (SRA) improves cross-branch character consistency by filtering irrelevant interactions and emphasizing high-correlation token pairs. 
While the full attention variant achieves slightly higher prompt fidelity due to its unrestricted receptive field, SRA reduces unnecessary cross-panel interactions and keeps long-context generation tractable while maintaining stronger coherence over extended sequences.

\begin{table}[t]
\centering
\caption{Ablation on attention configuration. SRA improves subject consistency, identity preservation, and visual quality while maintaining competitive prompt alignment.}
\resizebox{\linewidth}{!}{
\begin{tabular}{lcccc}
\toprule
\textbf{Method} & \textbf{VQAScore $\uparrow$} & \textbf{DreamSim Sim. $\uparrow$} & \textbf{FaceSim-Arc $\uparrow$} & \textbf{Aesthetic Score $\uparrow$} \\
\midrule
LCG (Full Attention) & \textbf{0.73} & 0.62 & 0.54 & 0.56 \\
LCG (Sparse Attention) & 0.71 & \textbf{0.75} & \textbf{0.66} & \textbf{0.62} \\
\bottomrule
\end{tabular}
}
\label{tab:ablation_attention}
\end{table}

\subsubsection{Routing Consistency Constraint}
Table~\ref{tab:ablation_rcc} evaluates the impact of the Routing Consistency Constraint (RCC) module. 
Incorporating RCC consistently improves both prompt alignment and character consistency, demonstrating its effectiveness in maintaining coherent features across panels.

\begin{table}[t]
\centering
\caption{Ablation on Routing Consistency Constraint (RCC). The results show that RCC effectively improves subject consistency and prompt alignment in long-context image generation.}
\resizebox{\linewidth}{!}{
\begin{tabular}{lcccc}
\toprule
\textbf{Method} & \textbf{VQAScore $\uparrow$} & \textbf{DreamSim Sim. $\uparrow$} & \textbf{FaceSim-Arc $\uparrow$} & \textbf{Aesthetic Score $\uparrow$} \\
\midrule
LCG (w/o RCC) & 0.67 & 0.69 & 0.60 & 0.59 \\
LCG (w/  RCC) & \textbf{0.71} & \textbf{0.75} & \textbf{0.66} & \textbf{0.62} \\
\bottomrule
\end{tabular}
}
\label{tab:ablation_rcc}
\end{table}

\subsubsection{Context Length}
\begin{table}[t]
\centering
\caption{Ablation on context length in LCG. Performance gradually decreases as the context length increases, while the 20-panel setting still maintains reasonable prompt alignment, subject consistency, facial identity preservation, and visual quality.}
\resizebox{\linewidth}{!}{
\begin{tabular}{lcccc}
\toprule
\textbf{Method} & \textbf{VQAScore $\uparrow$} & \textbf{DreamSim Sim. $\uparrow$} & \textbf{FaceSim-Arc $\uparrow$} & \textbf{Aesthetic Score $\uparrow$} \\
\midrule
LCG (6 panels)  & \textbf{0.71} & \textbf{0.75} & \textbf{0.66} & \textbf{0.62} \\
LCG (10 panels) & 0.70 & 0.72 & 0.63 & 0.59 \\
LCG (20 panels) & 0.68 & 0.69 & 0.59 & 0.58 \\
\bottomrule
\end{tabular}
}
\label{tab:ablation_context}
\end{table}

\begin{figure*}[p]
  \centering
  \includegraphics[width=0.95\textwidth]{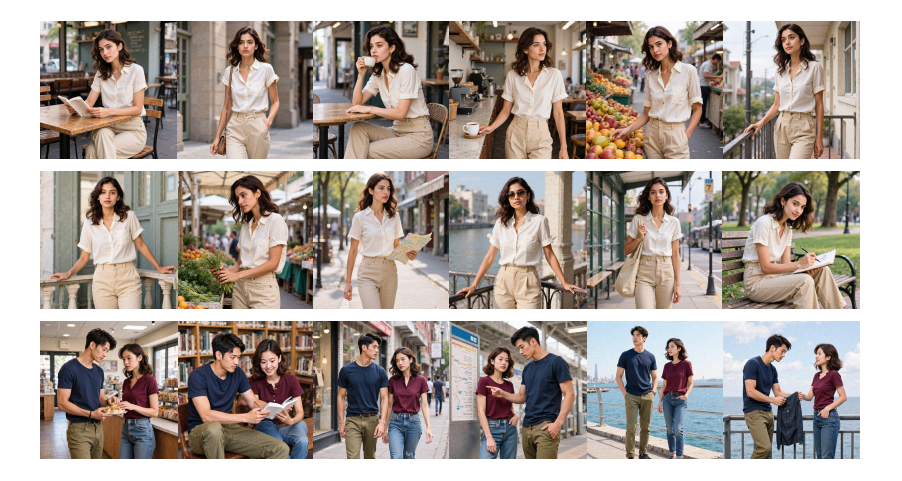}
  \caption{Additional long-context consistent image generation results. LCG preserves identity and appearance across single- and multi-character scenarios while following panel-specific prompts.}
  \Description{A full-width visual grid of multi-image generation examples. The generated panels show consistent character identity and clothing across different scenes and actions.}
  \label{fig:figures_only_main}

  \vspace{0.018\textheight}
  \includegraphics[width=0.74\textwidth]{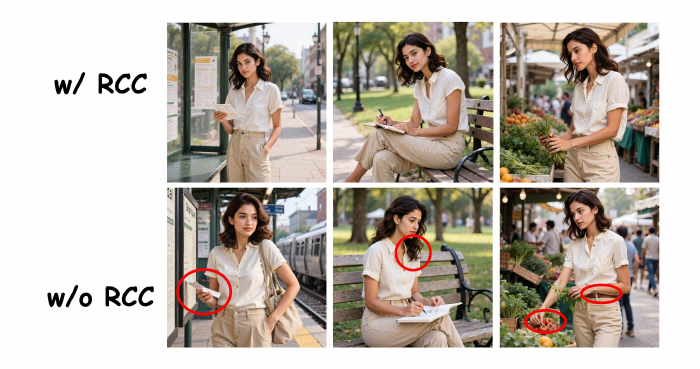}
  \caption{Qualitative comparison for the RCC ablation. Red circles highlight regions where removing RCC weakens prompt alignment, character consistency, or local visual stability.}
  \Description{A visual ablation comparing the full LCG model with a version without RCC. Red circles mark local regions where removing RCC causes weaker prompt adherence, identity drift, or artifacts.}
  \label{fig:rcc_visual}
\end{figure*}

To complement the quantitative results in \cref{tab:ablation_rcc}, \cref{fig:rcc_visual} provides visual comparisons between the full model and the variant without RCC. Removing RCC leads to weaker prompt realization, subtle identity drift, and occasional spurious artifacts, whereas RCC reduces these inconsistencies and stabilizes character appearance across scenes.

Table~\ref{tab:ablation_context} reports the performance of LCG under varying context lengths. As the number of simultaneously generated panels increases from 6 to 20, all four metrics show a gradual decline, reflecting the increased difficulty of long-context generation. Nevertheless, LCG retains competitive consistency at 20 panels, indicating that the proposed sparse cross-branch interaction remains effective under extended contexts.

\subsubsection{Efficiency Analysis}
\begin{figure*}[p]
  \centering
  \includegraphics[width=0.82\textwidth]{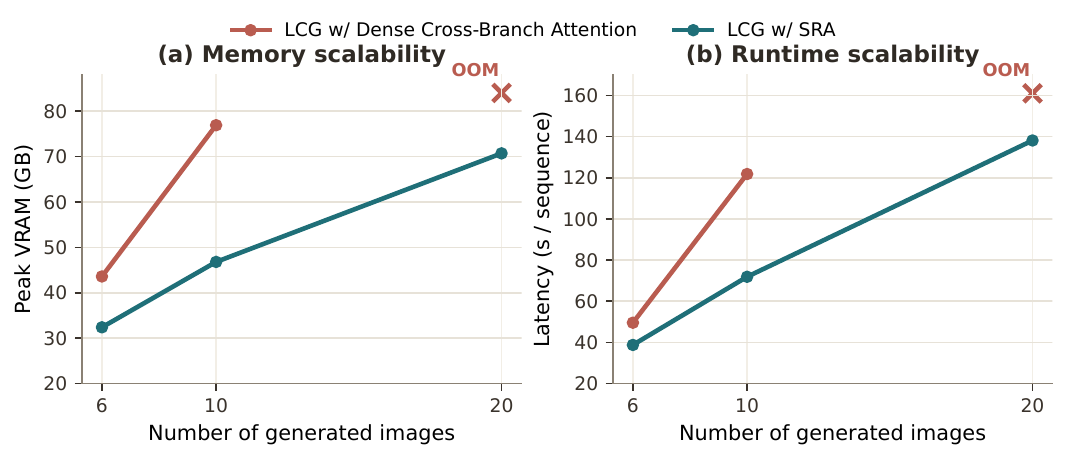}
  \caption{Scalability comparison between dense cross-branch attention and SRA on H800. SRA reduces peak memory usage and latency as the number of jointly generated images increases, while dense cross-branch attention runs out of memory at 20 panels under the same setting.}
  \Description{Two line plots comparing dense cross-branch attention and SRA. The first plot shows peak VRAM versus the number of generated images, and the second plot shows latency versus the number of generated images.}
  \label{fig:sra_efficiency}

  \vspace{0.03\textheight}
  \includegraphics[width=0.82\textwidth]{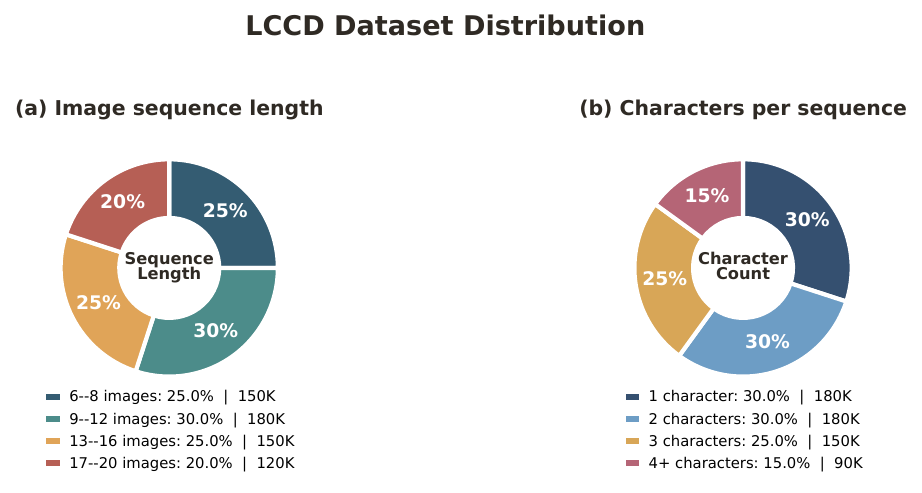}
  \caption{LCCD training-set distribution. Left: sequence length distribution over 6--20 generated images. Right: character-count distribution across training sequences.}
  \Description{Two donut charts summarizing the LCCD training-set distribution. The first chart shows sequence length groups, and the second chart shows the number of characters per sequence.}
  \label{fig:lccd_distribution}
\end{figure*}

We further evaluate the inference scalability of SRA by comparing it with a dense cross-branch attention variant under the same H800 setting. As shown in \cref{fig:sra_efficiency}, SRA consistently reduces peak VRAM and latency across different sequence lengths. The advantage becomes more pronounced as the number of jointly generated images increases: dense cross-branch attention remains feasible at 6 and 10 panels but runs out of memory at 20 panels, while SRA can still generate 20-panel sequences. These results support the design choice of using sparse relational routing for long-context multi-image generation.

\section{Conclusion}

We presented LCG, a framework for long-context consistent image generation. LCG combines Sparse Relational Attention (SRA) for scalable cross-branch interaction with the Routing Consistency Constraint (RCC) for more reliable identity and role alignment across jointly generated images. We also introduced LCCD, a large-scale synthetic dataset for training and evaluating long-context multi-image generation. Experiments show that LCG improves prompt alignment, identity consistency, and human preference over prior multi-image generation baselines. While LCG improves long-context consistency, maintaining fine-grained identity details remains more challenging as the sequence length and the number of interacting characters increase. Future work will explore stronger identity tracking and routing mechanisms for longer and more crowded multi-character sequences.

\bibliographystyle{ACM-Reference-Format}

\clearpage

\clearpage

\clearpage
\appendix
\section{Implementation Details}
\label{app:implementation}
\subsection{Model Architecture Design}
To help LCG distinguish multiple simultaneous generation branches, we exploit the temporal dimension of the 3D Rotary Position Embedding (3D RoPE). For each generation branch $b$ assigned to prompt $p_b$, we assign a fixed unique branch identifier $\tau_b$ and inject it into the temporal component of RoPE. This branch-level temporal signature gives each branch an explicit positional identity, helping reduce identity ambiguity across long-context generation. In the optional user-specified identity setting, clean identity images can be injected into selected branches using the same branch-level encoding, but such images are not required by the main multi-image text-to-image protocol.

In addition, practical observations indicate that distributing the Sparse Relational Attention (SRA) modules across the Flux.1-dev backbone yields more stable generation behavior. Specifically, inserting three SRA modules uniformly within each Double Block and six modules within each Single Block leads to consistently reliable performance in both multi-image and single-image generation scenarios.

\subsection{Training Strategy and Objectives}
Our training pipeline is organized into two sequential stages. In Stage 1, the newly introduced SRA modules are optimized on the 300K single-character training sequences, enabling the model to acquire a basic identity-preserving generation capability. In Stage 2, training continues on the 300K multi-character training sequences, allowing the model to further specialize in role binding and multi-subject consistency. The test set contains 1K sequences, split into 500 single-character and 500 multi-character sequences.
\section{Details of the LCCD Construction}
\label{app:lccd_details}

In this section, we provide a comprehensive description of the construction pipeline for the Long-Context Consistency Dataset (LCCD), a large-scale synthetic dataset for long-context multi-subject generation. LCCD consists of 600K training sequences, split equally between 300K single-character and 300K multi-character scenarios, and a separate 1K test set, split equally between 500 single-character and 500 multi-character scenarios. Each sequence contains 6 to 20 semantically coherent images paired with detailed descriptive prompts.

\paragraph{Construction pipeline.}
LCCD is constructed in four steps. First, Gemini 3 generates character profiles with facial attributes, clothing, age range, and style descriptions. Second, we compose 6-to-20-image prompt sequences by combining each character profile with scene, action, camera, and lighting descriptions. Third, Seedream 4.0 synthesizes one image for each prompt. Finally, Gemini 3 performs automatic consistency screening, and Grounded-SAM extracts character-level masks for RCC supervision.

\subsection{Synthetic-vs-Real Data Rationale}
Compared with real movie frames or web videos, synthetic construction has two practical advantages. First, it reduces legal risks related to copyright, privacy, portrait rights, and identity consent for identifiable real subjects. Second, it provides explicit control over character diversity, appearance attributes, scene categories, actions, camera compositions, lighting conditions, and multi-character interactions. Although LCCD does not replace real-world visual narratives, its prompts are designed to cover rich and realistic visual situations, making it a controlled and photorealistic dataset for studying long-context consistency.

\subsection{Data Synthesis via Seedream 4.0}
We leverage the Seedream 4.0 text-to-image model as our primary synthesis engine, chosen for its strong aesthetic fidelity and text-to-image alignment capabilities. Our generation strategy is grounded in the observation that large-scale foundational models exhibit a certain distributional bias after training, where highly detailed and specific prompts tend to anchor the model's output toward a consistent visual distribution. By providing exhaustive character appearance descriptions alongside varied scene-specific prompts, we help maintain character consistency while supporting diverse scenes. 

\paragraph{Prompt template.}
Each scene prompt starts with concise style and camera instructions, followed by the character appearance, action, environment, and lighting description:
\textit{[visual style], [camera/shot type], [character identity and clothing], [action], [environment], [lighting].}
For example: ``Cinematic film style, medium shot, a young woman with wavy dark hair, wearing a white short-sleeved shirt and beige trousers, writing in a notebook on a wooden park bench, warm afternoon light.''

\subsection{Quality Control and Automated Refinement}
While prompt engineering provides a strong baseline for consistency, identity drift can still occur across extended sequences. Gemini 3 serves as an automatic consistency checker for generated sequences. A sequence is retained only if the characters preserve facial identity, hairstyle, clothing, and role assignment across images. Sequences with identity drift, role confusion, severe clothing changes, missing characters, or obvious visual artifacts are discarded. Subsequently, we utilize Grounded-SAM to perform automated instance segmentation, yielding precise character masks for each frame to supervise the Routing Consistency Constraint (RCC).

\paragraph{Gemini filtering prompt.}
Given a sequence of generated images and the corresponding character profiles, Gemini 3 is asked to determine whether each character preserves the same facial identity, hairstyle, clothing, and role assignment across all images. The evaluator flags identity drift, role confusion, severe clothing changes, missing characters, or obvious visual artifacts. We retain only sequences that pass this automatic consistency screening.

\subsection{Bias and Scope}
Because LCCD is synthetic and filtered with external models, it may inherit distributional biases from Seedream 4.0, Gemini 3, and Grounded-SAM. We therefore treat LCCD as a controlled synthetic dataset rather than an unbiased proxy for all real-world visual narratives.

\subsection{Qualitative Analysis and Visualizations}
\begin{figure*}[t]
    \centering
    \includegraphics[width=0.95\textwidth]{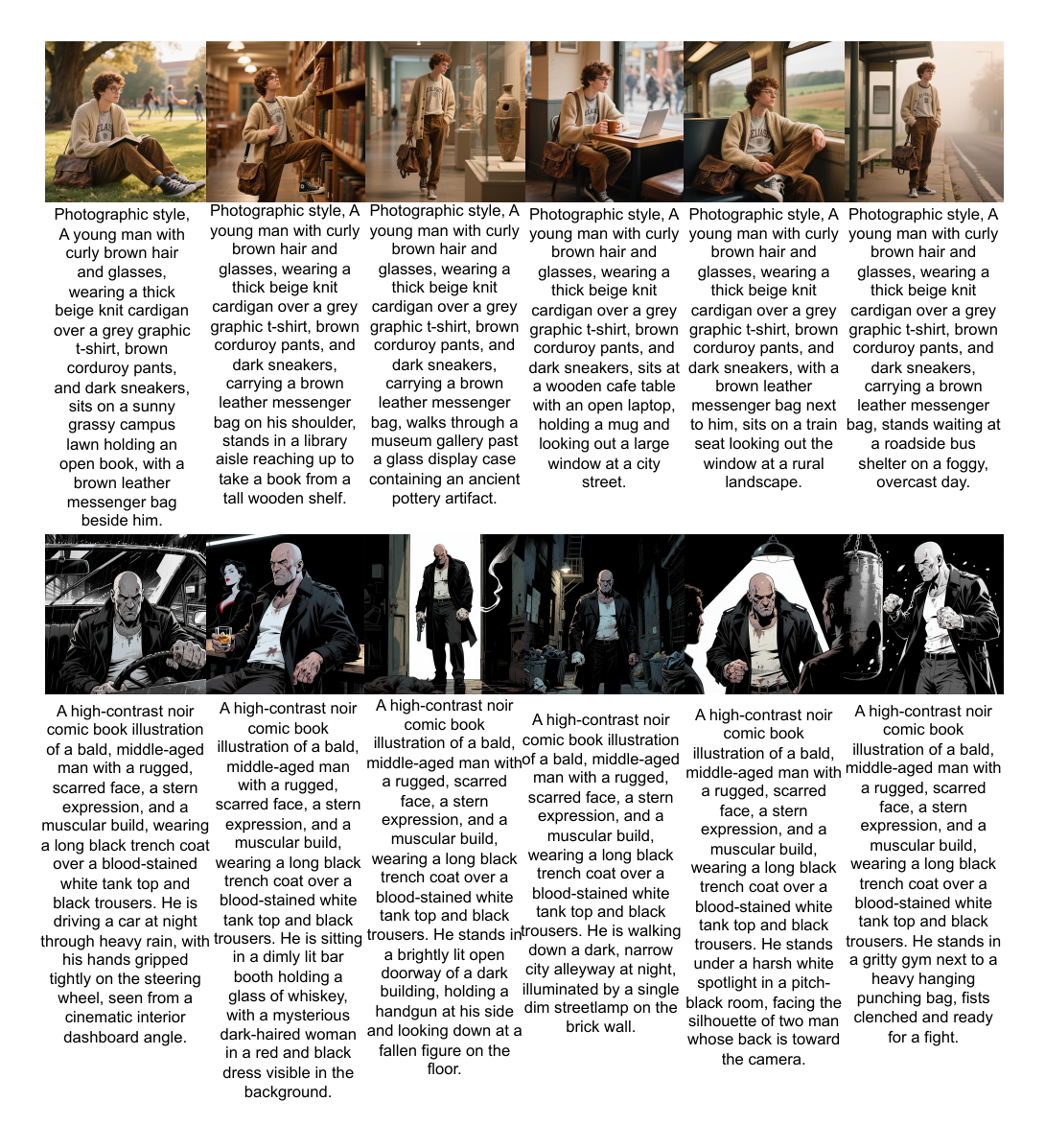} 
    \caption{Qualitative samples from the LCCD dataset (Part I). These 20-frame sequences show representative samples from our synthesis and filtering pipeline across diverse situational contexts.}
    \Description{A full-width grid of twenty-frame LCCD examples showing a character across diverse scenes while maintaining consistent visual identity and appearance.}
    \label{fig:sup9}
\end{figure*}

\begin{figure*}[t]
    \centering
    \includegraphics[width=0.95\textwidth]{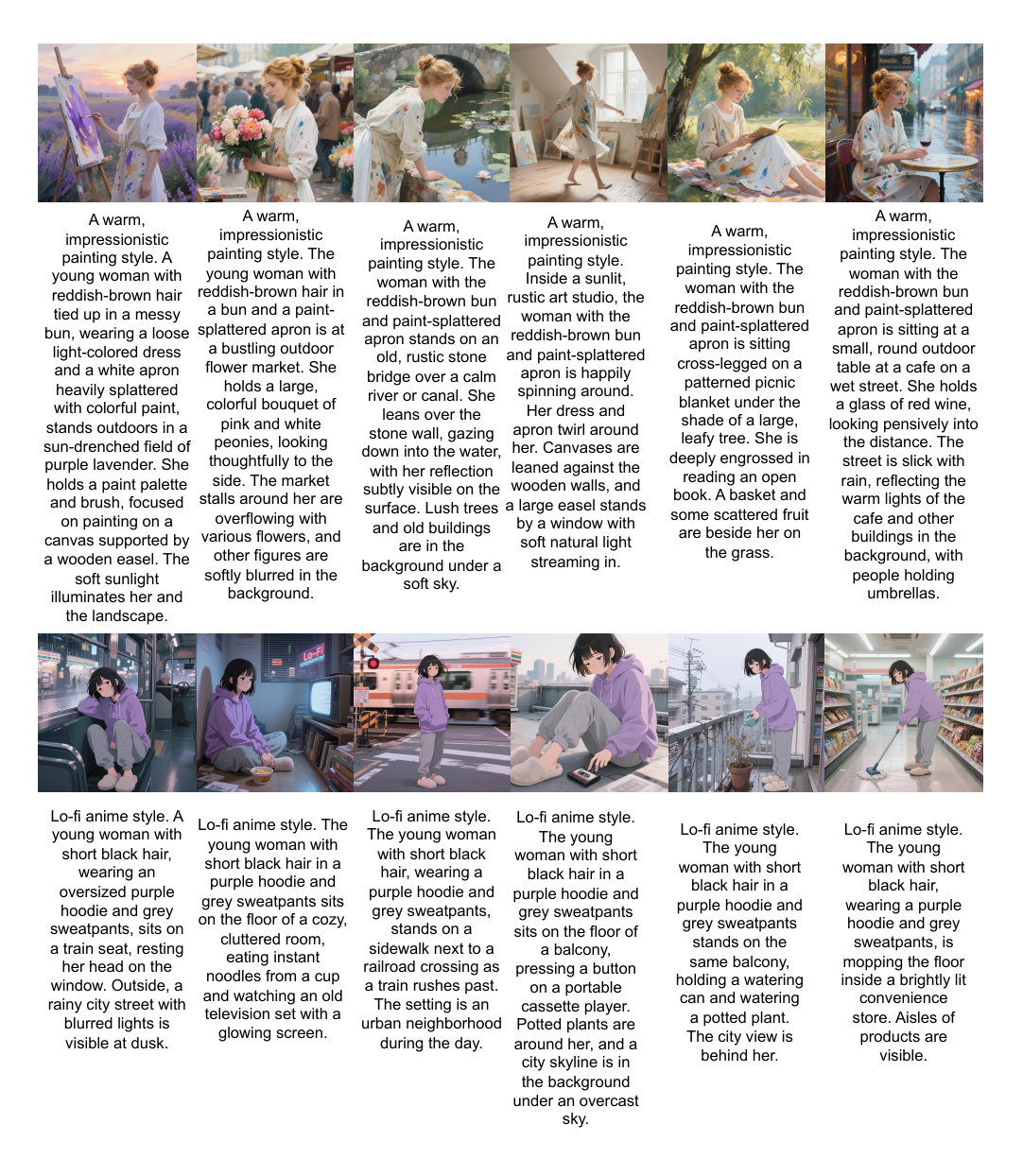} 
    \caption{Qualitative samples from the LCCD dataset (Part II). Additional examples demonstrating the high degree of character identity preservation and prompt alignment in the refined dataset.}
    \Description{A full-width grid of additional LCCD examples showing consistent character identity, clothing, and prompt alignment across multiple generated frames.}
    \label{fig:sup10}
\end{figure*}

The resulting LCCD dataset exhibits strong character consistency after the filtering pipeline and broad situational diversity. We provide representative qualitative samples across two consecutive pages to demonstrate this consistency. Figure~\ref{fig:sup9} showcases 20-frame sequences with persistent identity fidelity under varied environmental transitions. Furthermore, Figure~\ref{fig:sup10} provides additional examples highlighting the dataset's robust handling of complex character actions and lighting conditions.

\clearpage

\section{Additional Qualitative Results}
We provide additional qualitative visualizations for single-character and multi-character multi-image generation. These visual results demonstrate that our model consistently delivers strong prompt alignment and character-level consistency across diverse generation scenarios.

\subsection{Single-Character Multi-Image Generation}
\begin{figure*}[t]
    \centering
    \includegraphics[width=0.98\textwidth,height=0.94\textheight,keepaspectratio,trim=15bp 38bp 16bp 30bp,clip]{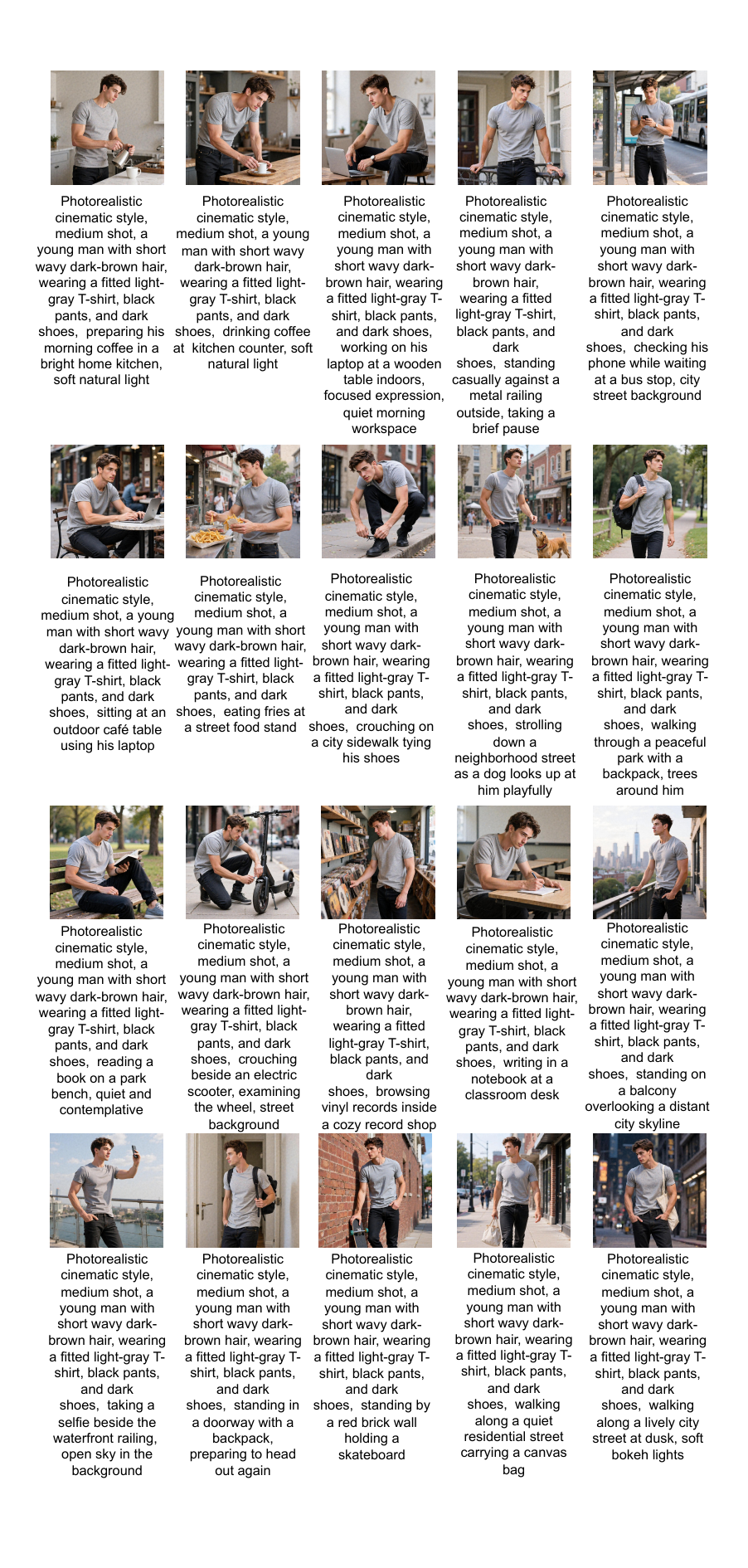}

    \caption{Single-character multi-image generation results of LCG.}
    \Description{Two rows of single-character multi-image generation results, showing one character across changing environments while preserving identity and clothing consistency.}
    \label{fig:sup1}
\end{figure*}

\begin{figure*}[t]
    \centering
    \includegraphics[width=0.98\textwidth,height=0.94\textheight,keepaspectratio,trim=15bp 28bp 16bp 28bp,clip]{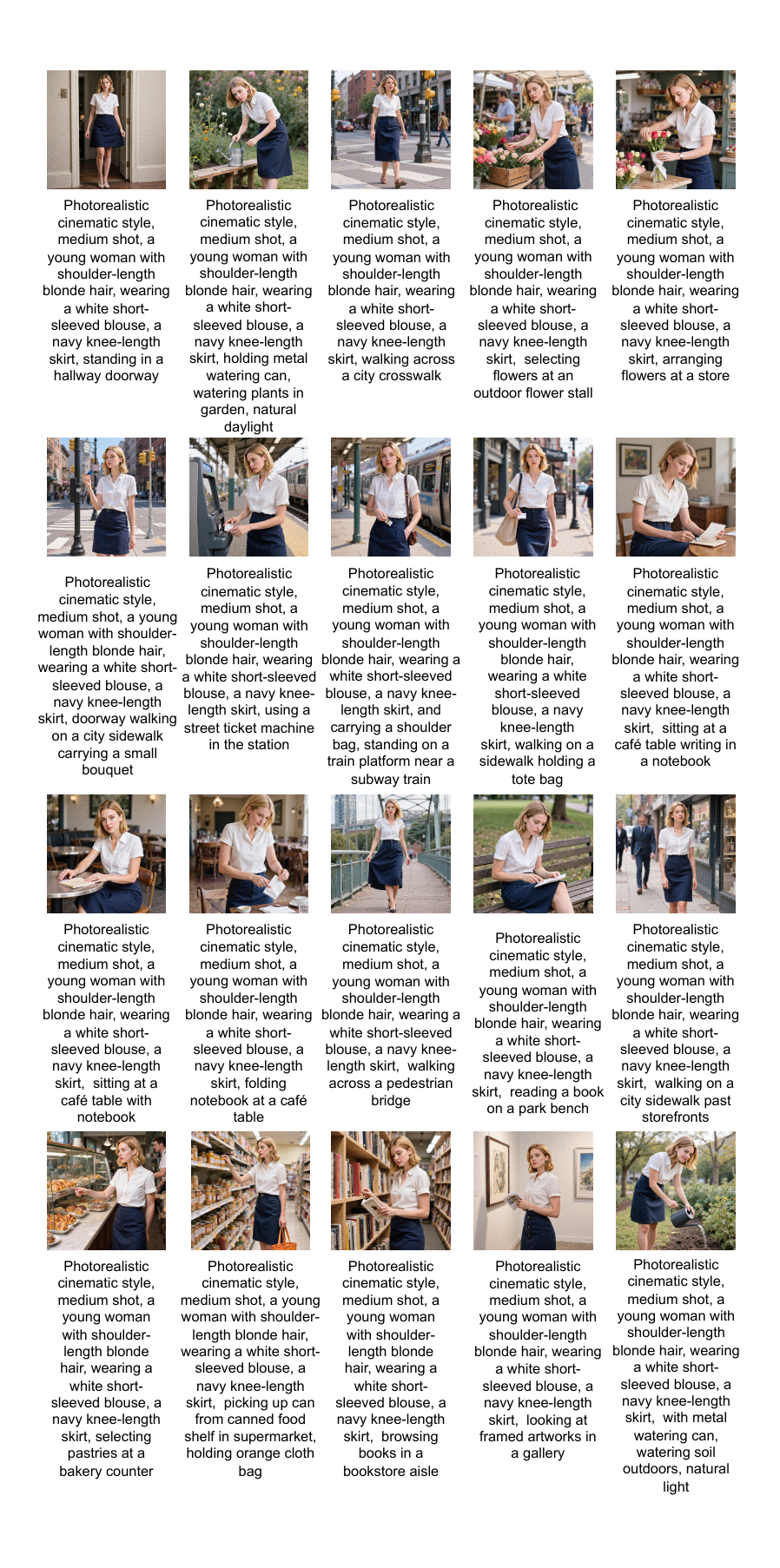}

    \caption{Additional single-character multi-image generation results of LCG.}
    \Description{Additional single-character multi-image generation examples showing consistent facial appearance and clothing across multiple generated panels.}
    \label{fig:sup2}
\end{figure*}

In this section, we present extended sequences of single-character generation. Figures~\ref{fig:sup1} and \ref{fig:sup2} demonstrate that LCG maintains high identity fidelity and clothing consistency even as the character transitions through diverse environmental contexts.

\subsection{Multi-Character Multi-Image Generation}
\begin{figure*}[t]
    \centering
    \includegraphics[width=0.98\textwidth,height=0.93\textheight,keepaspectratio,trim=15bp 20bp 16bp 20bp,clip]{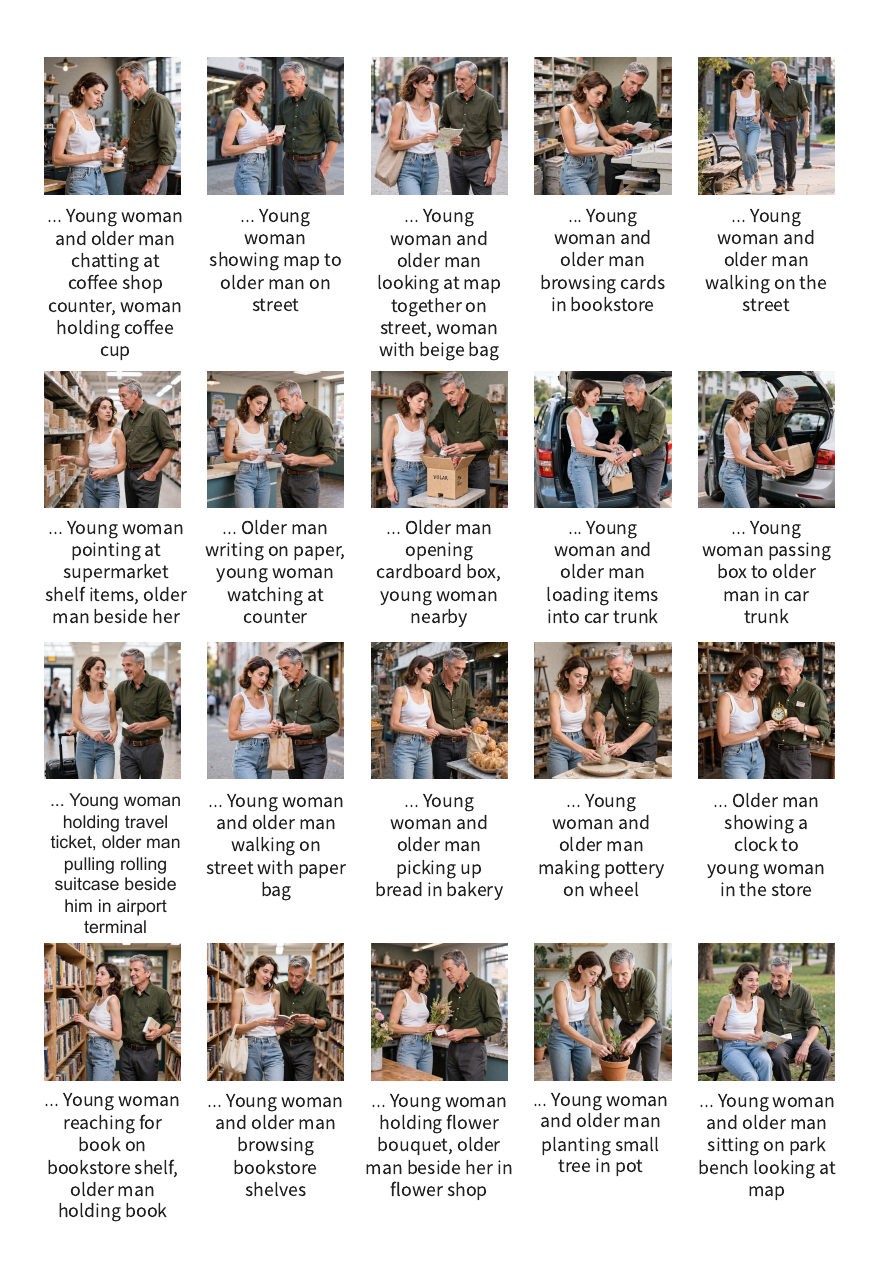}

    \caption{Multi-character multi-image generation results of LCG. For compactness, each prompt in the figure omits the shared prefix: ``Photorealistic cinematic style, medium shot, a young woman with long dark-brown hair wearing a white sleeveless top and blue jeans, and an older man with short gray hair wearing an olive-green shirt and dark trousers.''}
    \Description{Multi-character generation examples showing several characters interacting across multiple panels while retaining distinct visual identities.}
    \label{fig:sup3}
\end{figure*}

Figure~\ref{fig:sup3} showcases our model's capability in handling multi-subject interactions. By leveraging the Routing Consistency Constraint (RCC), LCG reduces identity confusion between different characters while preserving their individual traits throughout the storyboard.

\section{Additional Ablation Study Visualizations}
\label{app:ablation_visuals}

To complement the quantitative RCC ablation results in Section~\ref{sec:experiment}, we provide further visual evidence regarding the impact of our core components. As discussed in Section~\ref{sec:Method}, the Routing Consistency Constraint (RCC) is pivotal for maintaining identity and semantic stability across long-context sequences.

\begin{figure*}[t]
    \centering
    \includegraphics[width=0.95\linewidth]{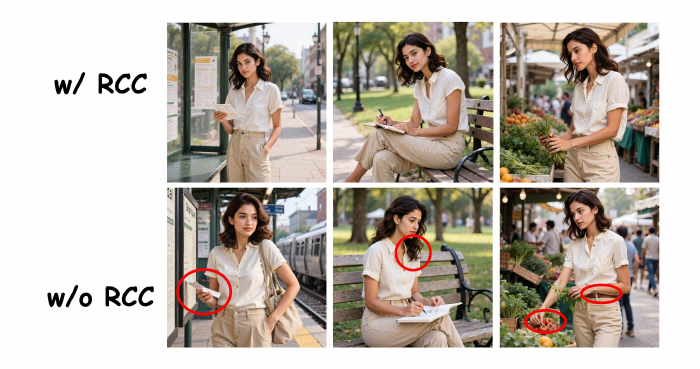}
    \caption{Visual comparison for the ablation of RCC. Red circles highlight regions showing degradation in prompt alignment and character consistency when RCC loss is removed. The corresponding prompts are:
    (1) A young woman with wavy dark hair, wearing a white short-sleeved shirt and beige trousers, standing on a train station platform holding a map;
    (2) The woman sitting on a wooden park bench while writing in a notebook;
    (3) The woman picking up fresh tomatoes at a vibrant outdoor market stall.}
    \Description{A side-by-side visual ablation comparing the full LCG model with a version without RCC. Red circles mark regions where removing RCC causes identity drift, artifacts, or weaker prompt adherence.}
    \label{fig:app_ablation_rcc}
\end{figure*}

Figure~\ref{fig:app_ablation_rcc} presents a side-by-side comparison between the full LCG model and a variant where the RCC loss is removed. As illustrated, the absence of RCC leads to noticeable degradation in both semantic adherence and character consistency. Typical failure patterns include:
\begin{itemize}
    \item \textbf{Identity Drift}: Subtle variations in facial structure across different frames (e.g., transitions between the station and the park).
    \item \textbf{Spurious Artifacts}: The emergence of unintended items, such as an inconsistent "belt" appearing only in specific frames.
    \item \textbf{Weakened Prompt Fidelity}: Incomplete realization of complex actions, such as "picking up tomatoes," where the alignment between local details and the descriptive prompt is noticeably weakened.
\end{itemize}

Incorporating RCC reduces these inconsistencies, helping the character's facial features and attire remain stable even under significant environmental transitions.

\end{document}